\documentclass[11pt,a4paper]{article}
\usepackage[hyperref]{emnlp2018}
\usepackage{times}
\usepackage{latexsym}
\usepackage{enumitem}
\usepackage{url}
\usepackage{paralist}

\usepackage{amsmath}
\usepackage{graphicx}
\graphicspath{{images/}}
\usepackage{tabularx}
\usepackage{multirow}

\usepackage{comment}
\usepackage{todonotes}

\aclfinalcopy % Uncomment this line for the final submission
\newcommand{\Note}[2]{} 
\newcommand{\SideNote}[2]{} 
\renewcommand{\Note}[2]{\todo[color=#1,size=\small, inline=true]{#2}} 
\renewcommand{\SideNote}[2]{\todo[color=#1,size=\small]{#2}} % 

\newcommand{\eat}[1]{\ignorespaces}

\title{Multi-task Learning for Universal Sentence Embeddings: A Thorough Evaluation using Transfer and Auxiliary Tasks}

\author{Wasi Uddin Ahmad\textsuperscript{$\dagger$}, Xueying Bai\textsuperscript{$\ast$}, Zhechao Huang\textsuperscript{\S}, Chao Jiang\textsuperscript{$\ast$}, Nanyun Peng\textsuperscript{$\star$}, Kai-Wei Chang\textsuperscript{$\dagger$}
\\
wasiahmad@ucla.edu, xb6cf@virginia.edu, huangzhechao1995@gmail.com \\
cj7an@virginia.edu, npeng@isi.edu, kwchang.cs@ucla.edu
\\ [3pt]
\textsuperscript{\S}Fudan University, \textsuperscript{$\ast$}University of Virginia \\
\textsuperscript{$\star$}University of Southern California, \textsuperscript{$\dagger$}University of California, Los Angeles
}

\date{}

\begin{document}
\maketitle

\begin{abstract}

Learning distributed sentence representations is one of the key challenges in natural language processing. Previous work demonstrated that a recurrent neural network (RNNs) based sentence encoder trained on a large collection of annotated natural language inference data, is efficient in the transfer learning to facilitate other related tasks. In this paper, we show that joint learning of multiple tasks results in better generalizable sentence representations by conducting extensive experiments and analysis comparing the multi-task and single-task learned sentence encoders. The quantitative analysis using auxiliary tasks show that multi-task learning helps to embed better semantic information in the sentence representations compared to single-task learning. In addition, we compare multi-task sentence encoders with contextualized word representations and show that combining both of them can further boost the performance of transfer learning.

\end{abstract}

\section{Introduction}
Learning distributional representations of words, phrases and sentences have long been interested in the natural language processing (NLP) community. 
In recent years, many successful stories showed that the learned representations are transferable to other tasks.
For example, many NLP ~\citep{zou2013bilingual, seo2016bidirectional,lee2017end,chen2017reading} and computer vision applications \citep{Venugopalan_2017_CVPR, teney2016graph} have successfully utilized word embeddings trained on a large corpus
% removed citation: bengio2003neural
\citep{mikolov2013distributed,pennington2014glove}. 
They are particularly beneficial when there is an insufficient amount of training examples to learn the word representations from scratch.
While the techniques of training word embedding are relatively matured and pre-trained word embeddings have been publicly available, learning high quality sentence representations is still a challenging problem.

\citet{conneau2017supervised} showed that a long short-term memory network (LSTM) \citep{hochreiter1997long} based sentence encoder trained on annotated corpus for natural language inference (NLI) task can capture useful features and is transferable to other tasks. 
By leveraging the annotations of a text classification task, the model forces the sentence encoder to embed semantic relations between sentences, and as a result, the encoder significantly outperforms existing unsupervised approaches \cite{kiros2015skip, hill2016learning} for learning sentence embedding. 
However, the NLI dataset is domain specific and relatively small in terms of vocabulary size\footnote{The largest NLI dataset, SNLI, only covers 42.7k unique tokens and extracted from image captions.}, limiting the performance of the universal sentence encoder. 
This leads to our research question: can we learn a better sentence encoder by leveraging more information from multiple different types of \emph{supervised} classification tasks?

In this paper, we study multi-task learning for training universal sentence representations.
We consider three large-scale text classification corpora, (1) Stanford Natural Language Inference (SNLI) \cite{bowman2015large}, (2) Multi-Genre NLI (Multi-NLI) \cite{williams2017broad}, and (3) Quora duplicate question pairs (Quora)\footnote{https://www.kaggle.com/quora/question-pairs-dataset}, which covers a variety of domains and two different tasks -- textual entailment and question paraphrasing.
We investigate two multi-task learning (MTL) frameworks, fully shared (FS) models and shared-private (SP) models to train a sentence encoder jointly on three tasks and examine their efficiency on 15 transfer tasks.
Our experiments show that sentence embeddings learned through multi-task learning outperform the single-task based universal sentence encoder \cite{conneau2017supervised}.

To achieve further improvement in transfer learning, we combine our proposed sentence encoder with recently proposed contextualized word vectors \cite{mccann2017learned,peters2018elmo} and show that the combined sentence encoding approach outperforms the state-of-the-art universal sentence embeddings learned using large-scale multi-task learning \cite{subramanian2018learning}.
In addition, we conduct quantitative analysis to evaluate the linguistic information captured by the multi-task learned sentence embeddings.

The contributions of the paper are summarized as follows.

% \NoteNP{Add contributions. Some ideas:}
% \begin{enumerate}
\smallskip
\begin{compactenum}
\setlength\itemsep{0.3em}
\item We study a multi-task learning framework to learn generalizable sentence representations which significantly outperform those learned from a single task on 7/10 transfer tasks and achieves comparable results on the rest. 
%, and empirically compare several  
%models considering different memory sharing strategies and training criteria for multi-task sentence representation learning.

% \item Our multi-task learned sentence representations significantly outperform those learned from a single task on 7/10 transfer tasks and achieves comparable results on the rest. 

\item We combine the multi-task learned sentence representations with contextualized word representations \cite{mccann2017learned, peters2018elmo} that yields new state-of-the-art results on 5/10 transfer tasks.

\item We analyze and evaluate the linguistic (both syntactic and semantic) information that is captured by the learned sentence representations using six auxiliary tasks. 
%, which can be useful for sentence embedding evaluation. %learned through single task and multi-task learning. 

% \item Our sentence encoders will be publicly available to facilitate research in related areas.
\end{compactenum}
% \end{enumerate}

\section{Related Work}

% kw: first talk about representation learning in general: e.g., representation of word and phrase.
Our work is closely related to sentence representation learning, transfer learning, and multi-task learning, because we propose to utilize multi-task learning to learn sentence representations that are helpful for transfer learning. We briefly review each of those areas in this section.

\smallskip
\noindent\textbf{$\bullet$ Sentence Representations Learning.}
Training complex networks to generate useful sentence representations has become a core component in many NLP applications. Recent works on learning distributional sentence representations such that they capture the syntactic and semantic regularities within sentences range from models that compose of word embeddings \cite{le2014distributed, arora2016simple, wieting2015towards} to more complex neural network architectures \cite{zhao2015self, wang2015learning, liu2016learning, lin2017structured}.
General purpose distributional sentence representations can be learned from a large collection of unlabeled text corpora.

\citet{kiros2015skip} proposed an {\em unsupervised} approach called SkipThought, by revising the skip-gram model \cite{mikolov2013distributed}  which are further improved by using layer normalization \cite{ba2016layer}.
% Skip-thought vectors help to achieve good results on 8 transfer tasks which are further improved by using layer normalization \cite{ba2016layer}.
\citet{hill2016learning} proposed two {\em unsupervised} objectives but they fell short of SkipThought.
Recently, \citet{logeswaran2018efficient} proposed to learn sentence representations by identifying the correct contextual sentences from a list of candidate sentences and showed improvement over SkipThought.
Unlike word embeddings, learning sentence representations in an unsupervised fashion lack the reasoning about semantic relationships between sentences. 
To learn universal sentence representations from {\em supervised} natural language inference data, \citet{conneau2017supervised} propose a BiLSTM with max pooling that yields the state-of-the-art in sentence encoding methods, outperforming unsupervised approaches like skip-thought vectors. 

A concurrent work \cite{subramanian2018learning} propose to build general purpose sentence encoder by learning from a joint objective of classification, machine translation, parse tree generation and unsupervised skip-thought tasks.
Compared to their approach, we employ multi-task learning to learn universal sentence encoders from multiple text classification tasks and combine them with existing contextualized word vectors \cite{mccann2017learned,peters2018elmo} for transfer learning.

% learning to learn general purpose sentence representations outperforming single-task based learned sentence representations. In this work, we also investigate {\em multi-task} learning of sentence embeddings, but focus on exploring the benefits of multi-task learning for yielding better sentence encoders.

% kw: Talk about transfer learning.
\smallskip
\noindent\textbf{$\bullet$ Transfer Learning.}
Transfer learning stores the knowledge gained from solving source tasks (usually with abundant annotated data), and apply it to other tasks (usually suffer from insufficient annotated data to train complex models) to combat the inadequate supervision problem. 
It has become prevalent in many computer vision applications \cite{sharif2014cnn, antol2015vqa} where image features were trained on ImageNet \cite{deng2009imagenet}, and NLP applications where word vectors \cite{pennington2014glove, mikolov2013distributed} were trained on large unlabeled corpora.
Despite the benefits of using pre-trained word embeddings, many NLP applications still suffer from lacking high quality generic sentence representations that are transferable to help other tasks. 
In this work, we investigate whether multi-task learning can help to obtain more robust sentence representations that transfer better. 

% talk about multi-task learning and their advantages on NLP applications.
\smallskip
\noindent\textbf{$\bullet$ Multi-task Learning.}
% The goal of multi-task learning is to improve generalization on the target task by leveraging  the domain-specific information contained in the training signals of related tasks~\citep{caruana1998Multitask}.
Multi-task learning has been successfully used in a wide-range of natural language processing (NLP) applications, including text classification \cite{liu2017adversarial}, machine translation \cite{luong2015multi}, sequence labeling \cite{rei2017semi}, sequence tagging \cite{peng2017multi}, dependency parsing \cite{peng2017deep} etc.
Recently, \citet{mou2016transferable} evaluated the utility of joint learning on sentence classification tasks to examine how much transferable neural networks are for NLP applications. However, they did not learn universal sentence encoders that can transfer information to unseen tasks as we do in this paper. 
% \NoteKW{The discussion in the following section is not very relevant to our paper, I think more importantly, you have to say why we are different from Liu.}
\citet{liu2016recurrent, zhang2017generalized} proposed multi-task learning architecture with different methods of sharing information, and can be further enhanced with an external memory \citep{liu2016deep} which is shared across the participant tasks.
% among the participating tasks. To further enhance, such models are augmented with external memory which is shared across the participant tasks \citep{liu2016deep}.
To facilitate scaling and transferring when a large number of tasks are involved, \citet{zhang2017multi} propose to embed labels by considering semantic correlations among tasks.

% As the first work on multi-task learning for sentence representation, we explore several different multi-task learning models  considering  different  memory  sharing strategies and training criteria to learn better sentence representations.

%Unlike these previous works, we intend to study two broad categories of the \emph{existing} multi-task learning architectures and utilize them to learn generic sentence representations that are transferable to other unseen tasks.
In this work, we study how to adapt the \emph{existing} multi-task learning architectures to learn generic sentence representations that are transferable.

\section{Multi-task Learning for Universal Sentence Encoder}
\label{sec:mtl_frameworks}

\begin{figure}[h]
\centering
\includegraphics[width=\linewidth]{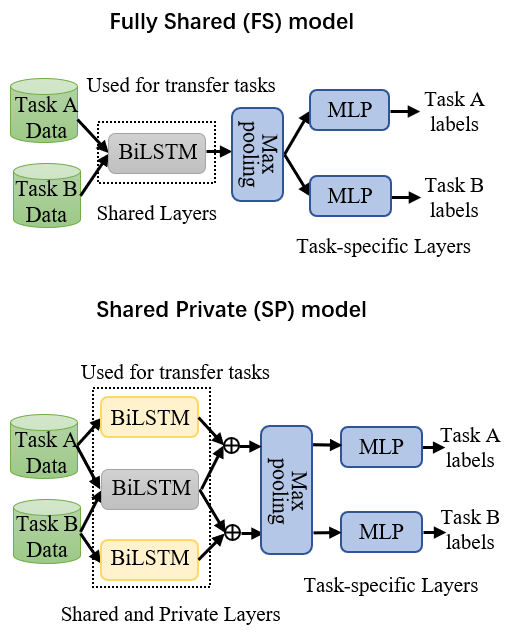}
\caption{Fully shared and shared private multi-task learning frameworks. $\oplus$ represents concatenation of sentence vectors. Yellow and gray boxes represent private and shared sentence encoders respectively. These encoders, after trained by multi-task learning, will be used for transfer learning.}
\label{mtl_frameworks}
\vspace{-3mm}
\end{figure}

While multi-task learning was shown efficient in many NLP tasks, its effectiveness in learning robust sentence representations that can generalize to unseen tasks is less studied. 
In this paper, we explore several existing multi-task learning frameworks for our goal of learning sentence representations. 
We hypothesize that with different training signals from various data and tasks, the learned sentence representations will be more robust and transferable. 
Our sentence encoders are trained on several sentence similarity or entailment datasets, and are employed to produce generic sentence representations for transfer tasks.
We study two hard parameter sharing frameworks\footnote{Two most common ways to perform multi-task 
learning are hard and soft parameter sharing. The former shares some parameters across all tasks while maintaining some task-specific parameters; the latter contains only task-specific parameters, but regularize the parameters of different tasks to be similar to encourage generalizable parameters. We confine our study to use hard parameter sharing models as they are widely used in the NLP community.} 
and explore two settings: the fully shared (FS) setting and the shared-private (SP) setting ~\citep{liu2017adversarial} (illustrated in figure \ref{mtl_frameworks}) to learn generalizable sentence representations.
We start by introducing the sentence encoder, which is the basic building block of the multi-task learning frameworks.

%%%%%%%%%%%%%%%%%%%%%%%%%%%%%%%%%%%%%%%%%%%%%%%%%%%%%%%%%

% \NoteKW{I think you need a short paragraph or subsection here to describe the  DNN architecture you use. You can simply say we follow the facebook paper to use the BiLSTM with max-pooling layer because this part is orthogonal to the multi-task learning approach you used. }
% solved - Wasi 
\subsection{Sentence Encoder}
A wide variety of neural network architectures can be used to convert sentences into fixed-size representations. We choose one layer bidirectional LSTM \citep{hochreiter1997long} with max pooling (BiLSTM-Max) sentence encoder as our basic building block for all multi-task learning architectures because it was found very effective in sentence encoding~\citep{conneau2017supervised}.
Formally, for a sentence with $T$ words $w=[w_{1}, w_{2},..., w_{T}]$, we have:
\begin{align}
\label{eqn:bilstm}
    \overrightarrow{h}_t & = LSTM(\overrightarrow{h}_{t-1}, w_{t}), \\ 
    \overleftarrow{h}_t  &= LSTM(\overleftarrow{h}_{t+1}, w_{t}), \\
    \mbox{and} \quad 
    h_t & = [\overrightarrow{h}_t, \overleftarrow{h}_t],
\end{align}
where $h_t \in R^{2d}$ is the $t$-th hidden vectors in BiLSTM, $d$ is the dimensionality of the LSTM hidden units.
To form a fixed-size vector representation of variable length sentences, the maximum value is selected over each dimension of the hidden units:
\begin{align}
    \label{eqn:maxpooling}
    s_{j} = \max\nolimits_{t \in [1,\ldots,T]} \ h_{j, t},\ j = 1, ..., d,
\end{align}
where $s_{j}$ is the $j$-th element of the sentence embedding $s$. Since we are coping with sentence entailment and similarity problems, there are always two sentences in one instance, denoted as $(s_{i1}, s_{i2})$, where $i$ indexes the instances. To get the representation of an instance, we follow \citet{liu2016learning} and define:
\begin{align*}
    s_i = & [s_{i1}, s_{i2}, s_{i1} - s_{i2}, s_{i1} \odot s_{i2}],
\end{align*}
where $\odot$ denotes the element-wise multiplication.

\subsection{Fully Shared (FS) models}
In fully-shared multi-task learning models, a single sentence encoder is shared across all tasks to learn generalizable representations.
The sentence representations obtained by following Eq.~\eqref{eqn:maxpooling}, are passed through task-specific
two-layer feed-forward neural networks to predict the task-specific target labels. 
\begin{align*}
    %s_i^k = & [s_{i1}^k, s_{i2}^k, s_{i1}^k - s_{i2}^k, s_{i1}^k \odot s_{i2}^k] \\
    \hat y_i^k = &\ \mbox{\it softmax}(W_2 \sigma(W_1 s_i^k + b_1)+b_2)
\end{align*}
where $i$ and $k$ indexes instances and tasks. $\sigma$ denotes element-wise sigmoid function.
We use cross-entropy loss. Thus the loss of each task is: 
% \NoteNP{Please add equations} - solved.
\begin{align*}
    \label{eq:cross_entropy_loss}
    L_{task}^k = - \sum\nolimits_{i=1}^{N_k} y_i^k \log(\hat y_i^k),
\end{align*}
where $y_i^k$ is the gold label of the $i$-th instance in task $k$.
The final multi-task loss $L_{task}$ is defined as a simple summation over the loss of each task.

%%%%%%%%%%%%%%%%%%%%%%%%%%%%%%%%%%%%%%%%%%%%%%%%%%%%%%%%%

\subsection{Shared-Private (SP) models}
Shared-private models contain shared encoders and private ones, which encourage the task-specific features being learned by the private encoders and generic features by the shared layers. 
In addition, each task has its own task-specific models to make the final prediction. 
We hypothesize that after proper multi-task training, the generic encoders are more useful than the private ones for transfer learning.
% Figure \ref{mtl_frameworks} illustrates a general architecture of the shared private models for multi-task learning.
In this study, we design one private BiLSTM-Max sentence encoder for each task, and one shared BiLSTM-Max encoder for all the tasks to capture generic features.
Sentence embeddings produced by private and shared encoders are concatenated to form the final sentence representations.
Formally, for any sentence in a given task $k$, its shared  representation $s^{k}_{s}$ and private representation $s^{k}_{p}$ can be computed using Eq. \eqref{eqn:bilstm} -- \eqref{eqn:maxpooling},
and the private and shared representations are concatenated to construct the sentence representations: $s^{k}=s^{k}_{s} \oplus s^{k}_{p}$.

\smallskip
\noindent\textbf{Adversarial Training.}
Ideally, we want the private encoders to learn only task-specific features, and the shared encoder to learn generic features.
To achieve this goal, we adopt the adversarial training strategy proposed by \citet{liu2017adversarial} to introduce a discriminator on top of the shared BiLSTM-Max sentence encoder.
The goal of the discriminator, $D$ is to identify which task an encoded sentence $s^k$ comes from, and the adversarial training requires the shared sentence encoder to generate representations that can ``fool'' the discriminator. In this way, the shared encoder is forced not to carry task-related information.
The discriminator is defined as,
\begin{align*}
    D(s^k) = \mbox{\it softmax}(W s^k + b),
\end{align*}
where $W \in R^{d \times d}$ and $b \in R^d$ are model parameters.
Optimizing the adversarial loss,
\begin{equation*}
    L_{adv} = \min_{\theta_{E}} \Bigg(\max_{\theta_{D}}\big(\sum_{k=1}^{K}\sum_{i=1}^{N_k} d_{i}^{k}\log[D(E(w))]\big)\Bigg)
\end{equation*}
has two competing goals: the discriminator tries to maximize the classification accuracy (inside the parentheses), and the sentence encoder tries to confuse it (and thus minimize the classification accuracy).
$E$ and $D$ represents the shared sentence encoder and the discriminator respectively and $\theta_E$ and $\theta_D$ are the model parameters of $E$ and $D$. $d_i^{k}$ denotes the ground-truth label indicating the type of the current task.
To encourage the shared and private encoders to capture different aspects of the sentences,
the following term is added.
\begin{align*}
    L_{diff} = \sum\nolimits_{k=1}^{K} \left \|  H_{s}^{k^\top} H_{p}^{k} \right \|_{F}^{2}.
\end{align*}
where $\left\|\cdot\right\|_{F}^{2}$ is the squared Frobenius norm. 
Here, $H_{s}^{k}$ and $H_{p}^{k}$ are matrices where rows are the hidden vectors (see Eq. (1)) generated by the shared and private encoders given an input sentence of task $k$.
The final loss function is a weighted combination of three parts:
\begin{align*}
    L & = L_{multi-task}+\beta L_{adv} + \gamma L_{diff},
\end{align*}
where $\beta$ and $\gamma$ are hyper-parameters, $L_{multi-task}$ refers to a simple summation over the cross entropy loss for each task.
We tune $\beta$ and $\gamma$ in the range $[0.001, 0.005, 0.01, 0.05, 0.1, 0.5]$ and found different combinations perform best for different models. 
Table \ref{table:param_tuning} indicates the best $\beta$ and $\gamma$ values for different multi-task learning settings.

%%%%%%%%%%%%%%%%%%%%%%%%%%%%%%%%%%%%%%%%%%%%%%%%%%%%%%%%%

% add the parameter tuning table
%%%%%%%%%%%%%%%%%%%%%%%%%%%%%%%%%%%%%%%%%%%%%%%%%%%%%%%%%%%%%%%%
\begin{table}[t]
\centering
% \resizebox{\linewidth}{!}{%
\small
\begin{tabular}{l|l|l}
\hline
Tasks & $\beta$ & $\gamma$ \\ 
\hline
DupQuD and SNLI & 0.01 & 0.05 \\
SNLI and Multi-NLI & 0.005 & 0.001 \\
DupQuD and AllNLI & 0.01 & 0.05 \\
DupQuD, SNLI and Multi-NLI & 0.005 & 0.001 \\
\hline
\end{tabular}
% }
\caption{Best $\beta$ and $\gamma$ values for adversarial shared private model on different set of tasks.}
% }
\label{table:param_tuning}
\vspace{-2mm}
\end{table}

%%%%%%%%%%%%%%%%%%%%%%%%%%%%%%%%%%%%%%%%%%%%%%%%%%%%%%%%%%%%%%%%

%%%%%%%%%%%%%%%%%%%%%%%%%%%%%%%%%%%%%%%%%%%%%%%%%%%%%%%%%

\subsection{Hyper-parameter Tuning} 
% \NoteNP{I won't use ``details'' in the title... like an engineer rather than a scientist. ;)}
We trained the aforementioned multi-task learning frameworks on SNLI, Multi-NLI, and Quora datasets.
We carefully tune the parameters on the development set and report the testing performance with best parameters. 
% For multi-task learning, we consider the average loss over all development examples.
We use SGD with an initial learning rate of $0.1$ and a weight decay of $0.99$. 
At each epoch, we divide the learning rate by $5$ if the development accuracy decreases. 
We use mini-batches of size $128$ and training is stopped when the learning rate goes below the threshold of $10^{-5}$. 
For the task-specific classifier, we use a multi-layer perceptron with $1$ hidden-layer of $512$ hidden units. 
For the hidden size of BiLSTM, we consider the range $[256, 512, 1024, 2048]$ and found $2048$ results in best performance. 
We use $300$ dimensional GloVe word vectors \citep{pennington2014glove} trained on $840$ billion of tokens as fixed word embeddings.
%We implemented all the models in PyTorch.

\subsection{Multi-task Learning for Contextualized Word Vectors}
With recent success of contextualized word vectors \cite{peters2018elmo} in many downstream NLP application, we are interested in investigating the utility of contextualized word vectors in our transfer learning setting. To this end, we modify the shared private multi-task learning framework by replacing the max-pooling layer with a more sophisticated biattentive pooling technique \cite{mccann2017learned} to obtain contextualized word vectors.
% We partially adopt the biattention technique proposed in \cite{mccann2017learned}. 
% At first, the pair of sentences, $s_x$ and $s_y$ are converted to contextualized word vectors using task-specific and shared bidirectional LSTM. 
We concatenate the task-specific and shared contextualized vectors produced by the bidirectional LSTMs for a pair of sentences, $s_x$ and $s_y$ to form sentence matrices $X$ and $Y$.
Then we compute an affinity matrix $A = XY^T$. The affinity matrix is used to extract the attention weights which are multiplied to the contextualized word vectors to get context summaries.
\begin{align*}
    A_x = softmax(A) & \quad A_y = softmax(A^T) \\
     C_x = A_{x}^{T}X & \quad C_y = A_{y}^{T}Y
\end{align*}
We concatenate the original representations, their differences from the context summaries, and the element-wise products between originals and context summaries.
\begin{align*}
    X_{|y} = & [X; X - C_y; X \odot C_y] \\
    Y_{|x} = & [Y; Y - C_x; Y \odot C_x]
\end{align*}
Then we apply max, mean, min and self-attentive pooling to form the sentence representations.
\begin{align*}
    \tilde{s}_{x} = & [max(X_{|y}); mean(X_{|y}); min(X_{|y}); x_{self}] \\
    \tilde{s}_{y} = & [max(Y_{|x}); mean(Y_{|x}); min(Y_{|x}); y_{self}]
\end{align*}
Where $x_{self} = X_{|y}^{T}\beta_x$ and $y_{self} = Y_{|x}^{T}\beta_y$.  $\beta_{x}$, $\beta_{y}$ are computed as follows.
\begin{align*}
    \beta_{x} = softmax(w_1X_{|y}), \beta_{y} = softmax(w_2Y_{|x})
\end{align*}
% \begin{align*}
%     \vspace{-2mm}
%     x_{self} = X_{|y}^{T}\beta_x \quad y_{self} = Y_{|x}^{T}\beta_y
% \end{align*}
Finally the sentence representations $\tilde{s}_{x}$ and $\tilde{s}_{y}$ are concatenated and passed through a task-specific feed-forward neural networks to predict the target labels.
We train both sentence embeddings and contextualized vectors on the same datasets. 

\section{Experimental Setup}
We hypothesize that learning sentence representations through multi-task learning can capture generic information and perform better on transfer tasks than the ones trained on single task.
To test this hypothesis, we learn generic sentence representations using multi-task learning on three large-scale textual entailment and paraphrasing datasets and test their generalizability of the sentence encoder on 15 additional transfer tasks.
% A detailed description of the source and transfer tasks are presented in table 1 in the appendix.
% ~\ref{table:data_stat}.
In addition, we perform a quantitative analysis using six auxiliary tasks to show what linguistic information is captured by the sentence representations.

%%%%%%%%%%%%%%%%%%%%%%%%%%%%%%%%%%%%%%%%%%%%%%%%%%%%%%%%%
%%%%%%%%%%%%%%%%%%%%%%%%%%%%%%%%%%%%%%%%%%%%%%%%%%%%%%%%%%%%%%%%
\begin{table}[ht]
\centering
\resizebox{\linewidth}{!}{%
\begin{tabular}{@{}l@{} c@{\hskip 0.1in} c@{\hskip 0.1in} c@{}}
\hline
Model Type & DupQuD & SNLI & Multi-NLI \\
\hline\hline 
\multicolumn{4}{l}{1. In domain single task sentence representation learning} \\ 
1.1 BiLSTM-Max & 86.7 & 84.5 & 70.8/69.8  \\
\hline \hline
\multicolumn{4}{l}{2. Multi-task sentence representation learning} \\ 
\hline
\multicolumn{4}{l}{2.1. 3-datasets and 2-tasks (DupQuD and SNLI)} \\
% \hline
2.1.1. Fully Shared & 86.6 & 84.8 & x \\
2.1.2. Shared Private & 86.8 & 84.7 & x  \\ 
2.1.3. Adversarial Shared Private & \underline{\textbf{87.0}} & \underline{84.9} & x    \\ 
\hline
\multicolumn{4}{l}{2.2. 3-datasets and 2-tasks (DupQuD and AllNLI)} \\ 
% \hline
2.2.1. Fully Shared & \underline{86.8} & \underline{84.8} & 70.0/68.9 \\
2.2.2. Shared Private & 86.0 & 84.7 & \underline{70.8/69.3}  \\ 
2.2.3. Adversarial Shared Private & 86.6 & 84.3 & 69.6/68.3   \\ 
\hline
% \hline
\multicolumn{4}{l}{2.3. 3-datasets and 3-tasks (DupQuD, SNLI and Multi-NLI)} \\ 
% \hline
2.3.1. Fully Shared & 85.9 & 84.3 & 70.1/69.6 \\
2.3.2. Shared Private & \underline{\textbf{87.0}} & \underline{\bf 85.2} & \underline{\bf 71.0/70.1}  \\
2.3.3. Adversarial Shared Private & 86.3 & 84.7 & 71.0/70.1  \\ 
\hline 
\hline
\multicolumn{4}{l}{3. Multi-task contextualized word vectors learning} \\ 
% \hline
3.1 SNLI + Multi-NLI & x & 85.8 & 73.3/72.3  \\
3.2 SNLI + Multi-NLI + DupQuD & 87.1 & 86.1 & 73.8/72.8  \\
\hline
\end{tabular}
}
\caption{Test accuracy of source tasks (DupQuD, SNLI and Multi-NLI) obtained through various multi-task learning architectures. Underlined values indicate best performance among models trained on the same set of tasks and bold-faced values indicate best performance across the board (excluding block 3).
}
\label{table:source_task_comparison}
\vspace{-3mm}
\end{table}
%%%%%%%%%%%%%%%%%%%%%%%%%%%%%%%%%%%%%%%%%%%%%%%%%%%%%%%%%%%%%%%%
%%%%%%%%%%%%%%%%%%%%%%%%%%%%%%%%%%%%%%%%%%%%%%%%%%%%%%%%%

%%%%%%%%%%%%%%%%%%%%%%%%%%%%%%%%%%%%%%%%%%%%%%%%%%%%%%%%%
\smallskip
\noindent\textbf{Source tasks.} The first source task is natural language inference (NLI) which determines whether a natural language hypothesis can be inferred from a natural language premise. 
We consider the SNLI \cite{bowman2015large} and the Multi-Genre NLI (Multi-NLI) \cite{williams2017broad} which consist of 
English sentence pairs, manually labeled with one of the three categories: entailment, contradiction and neutral.
% The SNLI dataset is constructed from image captions while the Multi-NLI dataset includes a more diverse range of text, as well as a test set for cross-genre transfer evaluation. 
Following \citet{conneau2017supervised}, we also conduct experiments that combine SNLI and Multi-NLI datasets, which is denoted as AllNLI.
The second task is the duplicate question detection (DupQuD) task based on a dataset of 404k question pairs released by Quora. We split the Quora dataset as that in \citet{wang2017bilateral}.

%%%%%%%%%%%%%%%%%%%%%%%%%%%%%%%%%%%%%%%%
%%%%%%%%%%%%%%%%%%%%%%%%%%%%%%%%%%%%%%%%%%%%%%%%%%%%%%%%%%%%%%%%
\begin{table}[h]
\centering
\resizebox{\linewidth}{!}{%
\small
\begin{tabular}{l|r|r|p{1.8cm}|c} %|p{8.5cm}}
\hline
Name & \multicolumn{1}{r|}{N} & \multicolumn{1}{r|}{V} & Task & C \\%& examples \\ 
\hline 
\hline
\multicolumn{4}{l}{Binary and multi-class classification tasks} \\ \hline
MR	    & 11k & 20.3k & sentiment & 2 \\ % & ``Too slow for a younger crowd , too shallow for an older one.'' (neg) \\
CR	    & 4k & 5.7k & product review & 2 \\ % & ``We tried it out christmas night and it worked great.'' (pos) \\
SUBJ	& 10k & 22.6k & subj/obj & 2 \\ % & ``A movie that doesn’t aim too high, but doesn’t need to.'' (subj) \\
MPQA	& 11k & 6.2k & opinion  & 2 \\ % & ``would like to tell''; (pos) \\
SST 	& 70k & 17.5k & sentiment  & 2 \\ % & ``the emotions are raw and will strike a nerve with anyone [..]'' (pos) \\
TREC	& 6k & 9.7k & question-type & 6 \\ % & ``What are the twin cities?'' (LOC:city) \\
\hline 
\hline
\multicolumn{4}{l}{Recognizing textual entailment tasks} \\ \hline
SNLI$^{\dag}$  & 560k & 42.7k &  entailment & 3 \\ % & ``A soccer game with multiple males playing.''; ``Some men are playing a sport'' (entailment) \\
Multi-NLI$^{\dag}$ & 433k & 102.7k &  entailment & 3 \\ % & ``The Old One always comforted Ca\'daan, except today.''; ``Ca\'daan knew the Old One very well.'' (neutral) \\
SICK-E 	& 10k & 2.4k &  entailment & 3 \\ % & ``A man is typing on a machine used
% for stenography''; ``The man isn’t operating a stenograph'' (contradiction) \\ 
\hline 
\hline
\multicolumn{4}{l}{Paraphrase identification tasks} \\ \hline
DupQuD$^{\dag}$ & 404k & 127.5k &  paraphrasing & 2 \\% & ``How do you start a bakery?''; ``How can one start a bakery business?'' (duplicate) \\
MRPC 	& 5.8k & 19.5k &  paraphrasing & 2 \\%& ``Ballmer has been vocal in the past warning that Linux is a threat to Microsoft.''; ``In the memo, Ballmer reiterated the open-source threat to Microsoft.'' (non-paraphrase) \\
\hline 
\hline
\multicolumn{4}{l}{Semantic textual similarity tasks} \\ \hline
SICK-R 	& 10k & 2.4k &  text similarity & 0 -- 5 \\%& ``A man is singing a song and playing the guitar''; ``”A man is opening a package that contains headphones”'' (1.6) \\
STSB  & 8.6k & 15.9k & text similarity & 5 \\
STS-12 	& 399 & 735 &  text similarity & 0 -- 5\\
STS-13 	& 561 & 1.6k &  text similarity & 0 -- 5\\
STS-14 	& 750 & 3.8k &  text similarity & 0 -- 5\\% & ``Liquid ammonia leak kills 15 in Shanghai''; ``Liquid ammonia leak kills at least 15 in Shanghai'' (4.6) \\ 3.7k in google sheet
STS-15 	& 750 & 1.3k &  text similarity & 0 -- 5\\
STS-16 	& 209 & 868 & text similarity & 0 -- 5\\
\hline
\end{tabular}
}
\caption{Statistics of the datasets for multi-task learning and the transfer tasks. N is the number of samples, V is the vocabulary size, and C is the number of classes or score range. $^{\dag}$ denotes the datasets that are used in multi-task learning.
}
\label{table:data_stat}
% \vspace{-1mm}
\end{table}

%%%%%%%%%%%%%%%%%%%%%%%%%%%%%%%%%%%%%%%%%%%%%%%%%%%%%%%%%%%%%%%%
%%%%%%%%%%%%%%%%%%%%%%%%%%%%%%%%%%%%%%%%

\begin{table}[ht]
\centering
\small
\begin{tabular}{l|c@{\hskip 0.1in} c@{\hskip 0.1in}| c@{\hskip 0.1in} c@{\hskip 0.1in}| c@{}}
\hline
& \multicolumn{2}{c|}{PTB} & \multicolumn{2}{c|}{ROCStories} & Senseval-3 \\ 
& N & $\mbox{S}_{Avg}$  & N & $\mbox{S}_{Avg}$ & N\\
\hline
Train   & 39,832 & 25.5 & 45,496 & 10.2 & 7,860 \\ 
Dev	    & 1,700 & 25.1 & 1,871 & 10.1 & - \\
Test    & 2,416 & 25.1 & 1,871 & 10.1 & 3,944 \\
\hline
\end{tabular}
\caption{Statistics of the auxiliary task datasets. For ROCStories, N indicates the number of stories and for PTB and Senseval-3, N indicates the number of sentences. $\mbox{S}_{Avg}$ is the average sentence length. For ROCStroies, spring 2016 version is used in the experiments.
}
\label{table:aux_data_stat}
\vspace{-3mm}
\end{table}

%%%%%%%%%%%%%%%%%%%%%%%%%%%%%%%%%%%%%%%%
% TABLE 3
%%%%%%%%%%%%%%%%%%%%%%%%%%%%%%%%%%%%%%%%%%%%%%%%%%%%%%%%%%%%%%%%
\begin{table*}[t]
\centering
\resizebox{\linewidth}{!}{%
% \begin{tabular}{@{}l@{\hskip 0.2in} c@{\hskip 0.2in} c@{\hskip 0.2in} c@{\hskip 0.15in} c@{\hskip 0.15in} c@{\hskip 0.15in} c@{\hskip 0.15in} c@{\hskip 0.1in} c@{\hskip 0.1in} c@{\hskip 0.1in} c@{}}
\begin{tabular}{@{}l@{\hskip 0.2in} c@{\hskip 0.15in} c@{\hskip 0.15in} c@{\hskip 0.08in} c@{\hskip 0.1in} c@{\hskip 0.1in} c@{\hskip 0.08in} c@{\hskip 0.08in} c@{\hskip 0.05in} c@{\hskip 0.1in} c@{}}
% \begin{tabular}{l c c c c c c c c c c}
\hline
Model Type & MR   & CR   & SUBJ & MPQA & SST  & TREC & SICK-R & SICK-E & MRPC & STSB  \\ 
\hline \hline
\multicolumn{11}{l}{1. Transfer learning from single-task sentence embeddings \cite{conneau2017supervised}} \\ 
% \hline
1.1. BiLSTM-Max (SNLI)    & 79.9 & 84.6 & 92.1 & 89.8 & 83.3 & 88.7 & 0.885 & 86.3  & 75.1/82.3 & x \\
1.2. BiLSTM-Max (AllNLI)   & 81.1 & 86.3 & 92.4	& 90.2 & 84.6 & 88.2 & 0.884 & 86.3 & 76.2/83.1 & x \\
\hline\hline
\multicolumn{11}{l}{2. Transfer learning from multi-task sentence embeddings} \\ 
% \hline
2.1. \citet{subramanian2018learning}   & 82.5 & 87.7 & 94.0 & \textbf{90.9} & 83.2 & 93.0 & \textbf{0.888} & \textbf{87.8}  & \textbf{78.6/84.4} & \textbf{78.9/78.6} \\
2.2. Shared Private (SP) & 81.6	& 86.9	& 93.9	& 89.2	& 84.4	& 90.4	& 0.883 & 85.9 &  76.5/83.3 & 77.3/76.9 \\
2.3. Adversarial Shared Private (ASP) & 82.0 & 86.3 & 93.8 & 89.4 & 84.1 & 92.2 & 0.884 & 87.0 & 77.2/83.6 & 77.8/77.3 \\
\hline \hline
\multicolumn{11}{l}{3. Transfer learning from contextualized word vectors} \\ 
% \hline
3.1. DupQuD, SNLI, and Multi-NLI & 80.9 & 83.2 & 91.2 & 88.8 & 84.6 & 91.6 & 0.760 & 83.0 & 74.0/71.8 & 66.6/67.8  \\
3.2. CoVe \cite{mccann2017learned} & 79.4	& 80.8	& 91.3	& 89.2	& 87.2 & 86.2	& 0.773 & 79.9	& 71.9/69.4 & 67.0/65.8  \\
3.3. ELMo \cite{peters2018elmo} & 82.6 & 85.7 & 94.3 & 89.3 & 87.5 & 94.0 & 0.785 & 79.3 & 75.1/73.1 & 71.6/70.6 \\
\hline \hline
\multicolumn{11}{l}{4. Transfer learning from multi-task sentence embeddings and contextualized word vectors} \\ 
4.1. CoVe + ELMo + Sent2Vec (SP) & 84.7 & 87.1 & 94.6 & 90.6 & 87.9 & 94.2 & 0.866 & 83.4 & 76.5/75.2 & 71.6/70.5 \\
4.2. CoVe + ELMo + Sent2Vec (ASP) & \textbf{84.8} & \textbf{88.4} & \textbf{94.8} & \textbf{90.9} & \textbf{88.5} & \textbf{94.8} & 0.865 & 84.1 & 77.9/77.4 & 71.8/71.1  \\
\hline
\end{tabular}
}
\caption{Transfer learning results for sentence representation learning using single-task, multi-task learning and contextualized word vectors. Bold-faced values denote the best results across the board. 
In block (2), the SP and ASP multi-task models are trained on three tasks (DupQuD, SNLI and Multi-NLI). 
In block (4), Sent2Vec refers to the shared encoder of the shared-private multi-task models. To form sentence representations from contextualized word vectors, we use a self-attentive pooling instead of max-pooling used to form MTL based representations.
% and in block (6), we conduct pair-wise comparisons between settings to highlight the differences.
}
\label{table:transfer_task_results}
\vspace{-1mm}
\end{table*}
%%%%%%%%%%%%%%%%%%%%%%%%%%%%%%%%%%%%%%%%

%%%%%%%%%%%%%%%%%%%%%%%%%%%%%%%%%%%%%%%%
% TABLE 4
%%%%%%%%%%%%%%%%%%%%%%%%%%%%%%%%%%%%%%%%%%%%%%%%%%%%%%%%%%%%%%%%
\begin{table}[ht!]
\centering
\resizebox{\columnwidth}{!}{%
% \scalebox{0.8}{
% \small
% \begin{tabular}{l c@{\hskip 0.3in} c@{\hskip 0.3in} c@{\hskip 0.3in} c@{\hskip 0.3in} c@{}}
\begin{tabular}{l c c c c c}
\hline
\multirow{2}{*}{Model Type} & \multicolumn{5}{c@{}}{Semantic Textual Similarity (STS)} \\
 & 2012 & 2013 & 2014 & 2015 & 2016 \\
\hline \hline
\multicolumn{6}{l}{Learning from single task} \\ 
% \hline
BiLSTM-Max (SNLI)    & 0.48 & 0.70 & 0.66 & \textbf{0.87} & 0.66 \\
BiLSTM-Max (AllNLI)  & \textbf{0.51} & 0.71 & \textbf{0.67} & 0.85 & 0.64 \\
\hline \hline
\multicolumn{6}{l}{Learning from multiple tasks (DupQuD, SNLI and Multi-NLI)} \\ 
% \hline
Shared Private & 0.45 & 0.71 & 0.63 & 0.79 & 0.66  \\ 
Adversarial Shared Private & 0.47 & \textbf{0.74} & 0.66 & 0.84 & \textbf{0.69} \\ 
\hline \hline
\multicolumn{6}{l}{Learning from contextualized word vectors} \\
% \hline
CoVe \cite{mccann2017learned} & 0.41 & 0.29 & 0.63 & 0.52 & 0.20  \\
ELMo \cite{peters2018elmo} & 0.45 & 0.38 & 0.62 & 0.65 & 0.20 \\
\hline 
\end{tabular}
}
\caption{STS12-16 transfer task results for different sentence representation learning. Bold-faced values indicate best performances. Sentence embeddings from the private and shared encoders are concatenated for the shared private MTL models in this experiment.
}
\label{table:unsup_task_comparison}
\vspace{-4mm}
\end{table}
%%%%%%%%%%%%%%%%%%%%%%%%%%%%%%%%%%%%%%%%%%%%%%%%%%%%%%%%%%%%%%%%

%%%%%%%%%%%%%%%%%%%%%%%%%%%%%%%%%%%%%%%%

%%%%%%%%%%%%%%%%%%%%%%%%%%%%%%%%%%%%%%%%%%%%%%%%%%%%%%%%%

\smallskip
\noindent\textbf{Transfer tasks.}
We evaluate the sentence encoders on 15 additional transfer tasks using the SentEval tool\footnote{https://github.com/facebookresearch/SentEval/}.
% The sentence evaluation tool released by \citet{conneau2017supervised} evaluates the quality of general purpose sentence embeddings on a broad set of transfer tasks.
Among them, six are text classification tasks for sentiment analysis (MR, SST), question-type (TREC), product reviews (CR), subjectivity/objectivity (SUBJ) and opinion polarity (MPQA). For these tasks, we report accuracy. 
% in table \ref{table:comparison}.
The SICK-E, SICK-R, MRPC, and STSB tasks are related to textual entailment and paraphrasing. 
% SICK-E task is very similar to SNLI, while MRPC is a paraphrase identification task which is very similar to the DupQuD task.
The tool provides accuracy and F1 score for MRPC but only accuracy for SICK-E.
Unlike SICK-E, SICK-R and STSB tasks predict a score between 0 and 5 given sentence pairs which indicate the level of textual similarity. 
For the SICK-R task, Pearson correlation is used as a measure and for the STSB task, the tool reports both Pearson and Spearman correlation. In addition, we report the Spearman correlation of STS12-16 tasks where two sentences are directly compared using cosine similarity based on the sentence representation.
Details of the transfer task datasets are provided in table \ref{table:data_stat}.

%%%%%%%%%%%%%%%%%%%%%%%%%%%%%%%%%%%%%%%%%%%%%%%%%%%%%%%%%
\smallskip
\noindent\textbf{Auxiliary tasks.}
To study what linguistic features are embedded into the sentence vectors, we conduct quantitative analysis using six auxiliary tasks. These six tasks are: sentence length prediction (8-class classification, \citet{adi2016fine}), part-of-speech tag prediction (46-class classification), word content and word order task (binary classification, \citet{adi2016fine}), word sense disambiguation (WSD), and sentence ordering.
Among these six tasks, we call the first four as syntactic auxiliary and the other two as semantic auxiliary tasks.
We conduct all the syntactic auxiliary experiments based on the Penn Treebank (PTB) \cite{marcus1993building} dataset.
For all the classification tasks, we use a two-layer MLP on top of the pretrained sentence encoders and report classification accuracy.

Following \citet{melamud2016context2vec}, we conduct the supervised WSD task using Senseval-3 lexical sample dataset \cite{mihalcea2004senseval} using the publicly available tool\footnote{https://github.com/orenmel/context2vec} and report F1 score.
The sentence ordering task~\cite{li2016neural}  refers to arranging a set of sentences into a coherent text.
We implement the end-to-end pointer network on top of the pretrained sentence encoders proposed in \cite{gong2016end} to evaluate the learned sentence representations.
We conduct the sentence ordering task based on the ROCStories \cite{mostafazadeh2016corpus} dataset which contains stories consisting of five sentences.
We shuffled the sentences and use the pointer network to arrange them in the correct order. 
We report pairwise metrics (PM) and longest sequence ratio (LSR) for the sentence ordering evaluation.
We refer the reader to \citet{gong2016end} for details of the evaluation metrics.
Details of the auxiliary task datasets is provided in table \ref{table:aux_data_stat}.

\section{Experimental Results}
We split our experiments into three parts. 
The first part examines the efficiency of multi-task learning for the source tasks; the second part evaluates the quality of the learned sentence encoders by using them to generate features for 15 different transfer tasks; and the third part quantitatively analyzes what syntactic and semantic information is captured by the trained sentence encoders.

\subsection{Evaluation on Source Tasks}
\label{subsec:eval_src_tasks}
In this section, we discuss the performance of both fully shared and shared-private multi-task learning (MTL) frameworks on different combinations of DupQuD, SNLI and Multi-NLI datasets as {\em source tasks}.
For the shared-private models, we concatenate the representations generated by shared and private (task-specific) encoders to form sentence embeddings.
The results can be found in table~\ref{table:source_task_comparison}.
We compare the performance of MTL with the models trained on single tasks as in \citet{conneau2017supervised}.
Table~\ref{table:source_task_comparison} shows that learning from multiple tasks performs better than learning from a single task.
Although contextualized word vectors learned through MTL performed best on the source tasks, in section \ref{subsec:transfer_eval}, we will show that the word vector encoders failed to outperform the sentence encoders on the transfer tasks.

%%%%%%%%%%%%%%%%%%%%%%%%%%%%%%%%%%%%%%%%%%%%%%%%%%%%%%%%%%%%%%%%

\begin{figure}[t]
\centering
% \vspace{-2mm}
\includegraphics[width=0.9\linewidth]{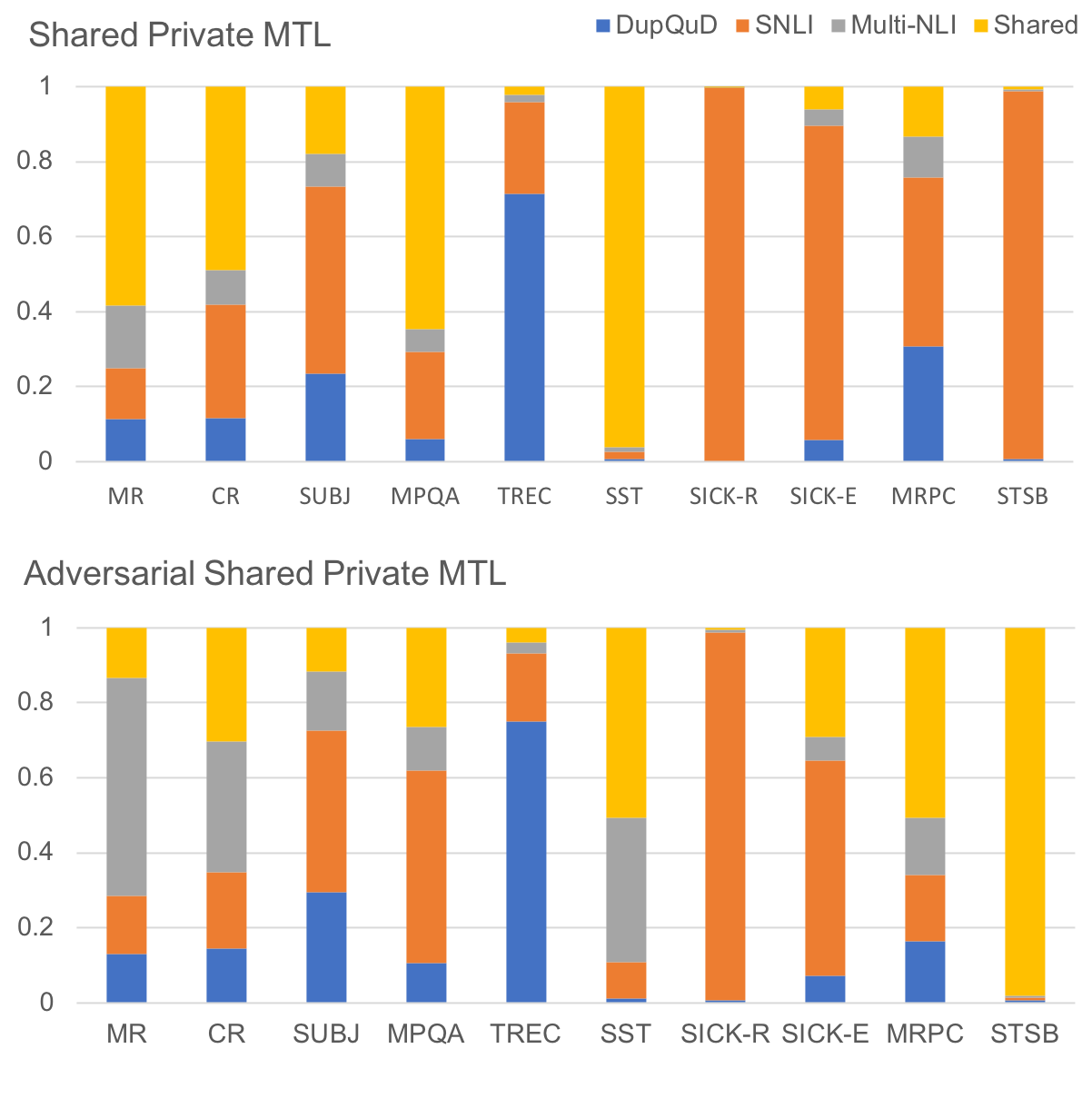}
\vspace{-5mm}
\caption{Weights learned by the transfer tasks for private (task-specific) and shared encoders of shared private multi-task models.}
\label{taskvsencoder}
% \vspace{-2mm}
\end{figure}

%%%%%%%%%%%%%%%%%%%%%%%%%%%%%%%%%%%%%%%%%%%%%%%%%%%%%%%%%%%%%%%%

%%%%%%%%%%%%%%%%%%%%%%%%%%%%%%%%%%%%%%%%%%%%%%%%%%%%%%%%%%%%%%%%

\begin{figure}[t]
\centering
% \vspace{-2mm}
\includegraphics[width=\linewidth]{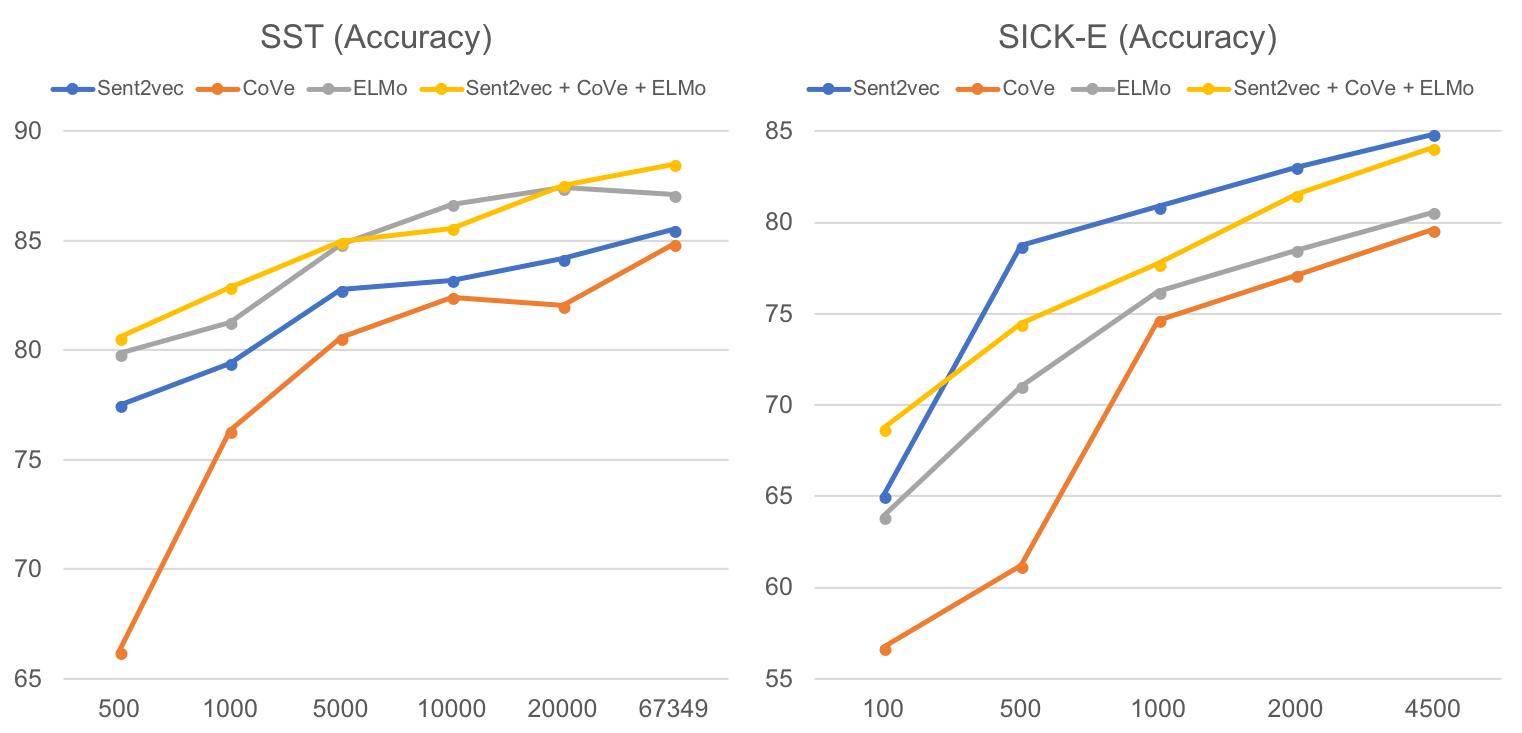}
% \vspace{-2mm}
\caption{Comparing performance of sentence representations and contextualized vectors on SST and SICK-E tasks as the training dataset size is varied (shown on x-axis). Sent2vec refers to the shared encoder in the shared-private MTL models.}
\label{sample_sicke}
\vspace{-2mm}
\end{figure}

%%%%%%%%%%%%%%%%%%%%%%%%%%%%%%%%%%%%%%%%%%%%%%%%%%%%%%%%%%%%%%%%

From table~\ref{table:source_task_comparison}, we see shared-private MTL outperforms fully shared MTL, especially when more tasks are involved in training.
This is intuitive because when more tasks are involved, it becomes harder to get good performance by modeling only the commonalities. So, having both task-specific and common features and allowing the final classifier to combine them is a sensible choice.
However, to our surprise, the adversarial training does not {\em always} excel on source tasks but in the next section, we will show that adversarial training boosts the transfer learning performance.

%%%%%%%%%%%%%%%%%%%%%%%%%%%%%%%%%%%%%%%%%%%%%%%%%%%%%%%%%%%%%%%%
%%%%%%%%%%%%%%%%%%%%%%%%%%%%%%%%%%%%%%%%%%%%%%%%%%%%%%%%%%%%%%%%
\begin{table}[ht]
\centering
% \resizebox{\linewidth}{!}{%
\small
\begin{tabular}{l|c|c}
\hline
\begin{tabular}{l} \textbf{Task} \end{tabular} & \begin{tabular}{@{}c@{}} \bf MTL $-$ STL \\ Same Data Size \\ More Tasks \end{tabular} & \begin{tabular}{@{}c@{}} \bf MTL $-$ STL \\ Larger Data Size \\ More Tasks \end{tabular} \\ 
\hline
MR	    & $+$0.1 & $+$1.0 \\ 
CR	    & $+$0.6 & $+$0.8 \\
SUBJ	& $+$1.6 & $+$1.3  \\
MPQA	& $-$0.7 & $-$0.6  \\
SST 	& $+$0.0 & $+$0.3  \\
TREC	& $+$1.7 & $+$2.1  \\
SICK-R	& $-$0.008 & $+$0.001  \\
SICK-E	& $-$1.5 & $+$0.6  \\ 
STS14 	& $+$0.7 & $+$1.4  \\
MRPC	& $-$0.003 & $+$0.0  \\
\hline
\end{tabular}
% }
\caption{The effect of more data/tasks on the sentence encoders for   transfer learning. The left column shows the differences between MTL and STL given equal amount of annotated data (from different task v.s. from the same task). The right column demonstrates the differences between MTL on two datasets and STL on one dataset (less data).}
%}
\label{table:wpb}
\vspace{-2mm}
\end{table}

%%%%%%%%%%%%%%%%%%%%%%%%%%%%%%%%%%%%%%%%%%%%%%%%%%%%%%%%%%%%%%%%
%%%%%%%%%%%%%%%%%%%%%%%%%%%%%%%%%%%%%%%%%%%%%%%%%%%%%%%%%%%%%%%%

\subsection{Evaluation on Transfer Tasks}
\label{subsec:transfer_eval}
In this part, we discuss the performances of sentence encoders learned by multi-task learning on 15 transfer tasks. Table~\ref{table:transfer_task_results} and \ref{table:unsup_task_comparison} summarizes the results. 
For the shared-private MTL models, we report the results of the concatenation of the sentence representations from both the shared and private encoders\footnote{We also compared the performance of private and shared encoders. The results can be found in appendix.}. 
%Similar to section \ref{subsec:eval_src_tasks}, w
% We analyze our findings by addressing the following research questions.
By comparing row 2.2--2.3 to 1.1--1.2 in table~\ref{table:transfer_task_results}, we see MTL based sentence representations outperform SNLI on 9 transfer tasks and 7 out of 10 for AllNLI (and comparable performances on the other 3 tasks).
% Row 4.1 shows that the improvements can be as much as 3.0\% and 2.0\% higher for TREC and MRPC tasks.
The results verify our hypothesis that learning from multiple tasks helps to capture more generalizable features that are suitable for transfer learning.

Comparing adversarial shared-private model to non-adversarial setting, we see improved performance on 4 out of 10 transfer tasks.
% Moreover, the improvement on TREC task is as much as 1.8\% higher (92.2\% vs. 90.4\%).
We observed that shared encoders perform in general better than private encoders, which confirms that shared encoders learn generic features that are more suitable for transfer learning. 
However, the concatenation of representations from both shared and private encoders results in better performance, which indicates that transfer tasks also get benefited from task-specific features because of the commonality with the source tasks.
To understand the relation between source and transfer tasks, we consider summing the weighted representations\footnote{In the experiment, we observed concatenating the private and shared sentence embeddings perform slightly better than a weighted summation with the logistic regression classifier.} produced by the private and shared encoders and the learned weights are shown in figure \ref{taskvsencoder}.
The learned weights indicate that with adversarial training the shared encoder gets less contaminated by private features and thus gets lower weight from most of the transfer tasks as opposed to non-adversarial training.
Also higher weights on the SNLI encoder assigned by most of the transfer tasks explains why sentence representations learned through single task learning based on the SNLI dataset gives competitive performance on most of the transfer tasks.

\begin{table*}[]
\small
\centering
\begin{tabular}{p{7.5cm}|p{7.5cm}}
\hline
\multicolumn{1}{c|}{\textbf{Source Sentence}} & \multicolumn{1}{c}{\textbf{Nearest Neighbor}} \\ \hline
% \{...\} the movies are at heart a form of fiction, like the \underline{play}, the novel, or the short story.                         & \{...\} worthy of being explored either in the novel or in the \underline{play} or in comedy and satire.                           \\ \hline
% \{...\} did the mound chores from West Palm Beach to \underline{play} the game before 767 paying customers \{...\}                   & The Texans have two more road games - at Buffalo and Houston - before they \underline{play} for the home folks again, \{...\}      \\ \hline
% \{...\} a small boy, who dropped this suddenly hot potato in a very playable \underline{lie}.                                        & \{...\} and the whole world is telling you that one was a \underline{lie}.                                                         \\ \hline
% He no longer had to \underline{lie} in his bunk all night, his eyes closed, pretending to sleep.                                     & I do it, lots of times - I like to \underline{lie} in a hammock at night, by myself, when it's all quiet.                          \\ \hline
 He may \underline{support} China (but he won't); he may break with China (which would be infernally difficult and perhaps disastrous), \{...\}                      & The Chinese, North Vietnamese and North Koreans, on the other hand, feel that, militarily, Russia is strong enough to \underline{support} them in the ``just wars of liberation'' \{...\}                  \\ \hline
\{...\} effected among project finance,  utilization of agricultural surpluses, and general balance of payments \underline{support}. & The local community maintains responsibility for the financial \underline{support} of its own library program, facilities, \{...\} \\ \hline
\end{tabular}
\caption{Nearest neighbors to ``support'' using sentence embeddings from shared-private MTL models. The source and nearest neighbor sentences are picked from the SemCor 3.0 dataset.}
\label{polysemous}
\end{table*}
%%%%%%%%%%%%%%%%%%%%%%%%%%%%%%%%%%%%%%%%%%%%%%%%%%%%%%%%%%%%%%%%
\begin{table*}[h]
\centering
% \resizebox{\linewidth}{!}{%
\small
\begin{tabular}{l c@{} c@{} c@{} c@{} c@{} c}
% \begin{tabular}{l@{\hskip 0.05in} c@{\hskip 0.05in} c@{\hskip 0.05in} c@{\hskip 0.05in} c@{\hskip 0.05in} c@{\hskip 0.05in} c@{}}
\hline
Model Type  & \begin{tabular}{c} Length \\ Prediction \end{tabular}  & \begin{tabular}{c} POS tag \\ Prediction \end{tabular} & \begin{tabular}{c} Word \\ Content \end{tabular}  & \begin{tabular}{c} Word \\ Order \end{tabular} & \begin{tabular}{c} WSD \end{tabular} & \begin{tabular}{c@{}} Sentence \\ Ordering \end{tabular}  \\ 
\hline
% Majority vote classifier  & -  & 84.23	& 75.33 & 57.04 \\ 
\hline
\multicolumn{7}{l}{Sentence representation learning from contextualized word vectors} \\ 
%\hline
CoVe \cite{mccann2017learned}   & \textbf{77.2} & 94.9 & 73.7 & \textbf{70.2} & 46.5 & 75.4/74.9 \\
ELMo \cite{peters2018elmo}   & 58.1 & \textbf{97.5} & 77.5 & 68.5 & \textbf{66.5} & \textbf{81.1/79.3} \\
\hline
\hline
\multicolumn{7}{l}{Sentence representation learning from single-task} \\ 
%\hline
BiLSTM-Max (SNLI)    & 72.9 & 96.8 & 79.4 &	65.5 & 39.0 & 79.0/77.3 \\ 
BiLSTM-Max (All-NLI)   & \underline{73.8} & 97.0 & \textbf{80.2} & 67.1 & 39.0 & 78.1/76.5 \\
\hline
\hline
\multicolumn{7}{l}{Sentence representation learning from three-tasks (DupQuD, SNLI and Multi-NLI)} \\ 
Shared Private & 66.9 & 96.6 & \underline{79.7} & \underline{69.7} & 45.6 & 78.6/77.2 \\
Adversarial Shared Private & 69.7 & \underline{97.1} & 79.1 & 68.3 & \underline{48.1} & \underline{79.3/77.7} \\
\hline
\end{tabular}
% }
\caption{Auxiliary task results for sentence representations learned from the best two single-task and multi-task models and contextualized word vectors. 
We only consider the shared encoder from the shared private multi-task models for auxiliary task experiments.
Bold-faced and underlined values denote the best and second best results across the board.
}
\label{table:aux_comparison}
\vspace{-3mm}
\end{table*}

%%%%%%%%%%%%%%%%%%%%%%%%%%%%%%%%%%%%%%%%%%%%%%%%%%%%%%%%%%%%%%%%

In table \ref{table:transfer_task_results}, row 3.3 demonstrates that ELMo outperforms MTL based sentence embeddings on MR, SUBJ, SST and TREC by a large margin. This motivates us to combine the representations learned by MTL with ELMo and CoVe\footnote{In our experiment, we found that CoVe, ELMo and MTL based representations gives better performance compared to a combination of ELMo and MTL based representations.} and from row 4.2 in table \ref{table:transfer_task_results}, we see a large improvement on 5 transfer tasks. In addition, we study the efficiency of CoVe, ELMo, MTL based representations and the combined encoding approach based on different dataset size using the SST and SICK-E tasks and the figure \ref{sample_sicke} summarizes our findings.

From table \ref{table:unsup_task_comparison}, we see that sentence embeddings learned from MTL and STL significantly outperforms CoVe and ELMo, demonstrating the necessity of generic sentence representations that can be directly used (without any training) in transfer tasks. 
However, MTL based sentence embeddings failed to outperform STL based ones on most of the STS12-16 tasks due to the similarity of the source and transfer tasks. For example, STS-16 contains sentence pairs on plagiarism detection which is similar to the DupQuD task and because of that MTL outperforms STL on this task.

When we compare MTL to STL for transfer learning, one fundamental question that arises is, does improvement in transfer learning via MTL only come because of having more annotated data? Comparing the performance of AllNLI in a single task setting and \{SNLI, Multi-NLI\} in the multi-task settings, we observe significant improvement in 7/10 tasks. In both settings, the amount of training data is the same. 
To verify the hypothesis that the improvements in transfer learning do not solely come from having more annotated data, we design an experiment that samples equal amount of data (225k training examples) from SNLI and DupQuD to match the size of full SNLI dataset.
We found 0.26\% average improvement in transfer tasks compared to single task learning (STL) on the SNLI dataset. 
With full SNLI and DupQuD dataset, we observe a larger (0.69\% on average) improvement in transfer tasks compared to STL on SNLI dataset.
Left column of table~\ref{table:wpb} shows that MTL is beneficial in this setting and the right column demonstrates that with additional data, MTL achieves larger gains.
% \footnote{more details are provided in the appendix.}.

%%%%%%%%%%%%%%%%%%%%%%%%%%%%%%%%%%%%%%%%%%%%%%%%%%%%%%%%%%%%%%%%%%%%%%%%%%%%%%%%%%%%%%%%%%%%%%%%%%

\subsection{Analysis using Auxiliary Tasks}
We hypothesize that the universal sentence encoders embed syntactic and semantic information of sentences so that they would be helpful for various transfer tasks.
To test this hypothesis and evaluate the quality of MTL based sentence embeddings, we employ six auxiliary tasks and compare with STL and contextualized vectors based sentence representations.
Table \ref{table:aux_comparison} demonstrates that sentence embeddings learned from CoVe and ELMo performs better but the MTL based sentence representations results in  competitive performance in majority of the auxiliary tasks. Particularly, the performance of MTL based sentence encoders in word sense disambiguation tasks shows that the multi-task learning helps to capture better linguistic information. Table \ref{polysemous} shows that the MTL based sentence embeddings is capable of disambiguating both the part of speech and word sense in the source sentence.

To compare the private and shared encoders in shared private MTL models, we conduct the auxiliary experiments on individual encoders. In our experiments, we found that the sentence encoder trained on Multi-NLI consistently outperforms other task-specific or shared encoders on the auxiliary tasks.
% Although we consider Multi-NLI as a single task, 
Multi-NLI dataset consists of sentence pairs from five different genres of spoken and written text and as a result, sentence encoders trained on Multi-NLI capture more diversified and rich syntactic and semantic features.
We hypothesize having annotated data covering a variety of \emph{domains} can better help multi-task learning.

\section{Conclusion}
In this paper, we investigate the effectiveness of multi-task learning (MTL) for training generalizable sentence representations by evaluating on both source tasks and 10 different transfer tasks.
Experiments on two categories of MTL frameworks demonstrate that multi-task learning outperforms single-task learning both on the source tasks and most of the transfer tasks.
In addition, we analyze what linguistic information is captured by the sentence representations.
In our future work, we will explore advanced techniques to model domains: in Multi-NLI, there are five different domains, which we do not explicitly model. We will study whether multi-task learning can be benefited from proper modeling of domains. 
% In addition, we will explore better models for multi-task learning to improve the semantic information captured by the universal sentence encoder.

% Use unnumbered third level headings for the acknowledgments. All acknowledgments, including those to funding agencies, go at the end of the paper.

% \newpage
% \small
\bibliography{emnlp2018}
\bibliographystyle{emnlp_natbib}

\appendix
\begin{table*}[ht]
\centering
% \resizebox{\linewidth}{!}{%
% \small
\begin{tabular}{l c c c c c c}
\hline
\multirow{2}{*}{Model Type} & \multicolumn{2}{c}{Quora} & \multicolumn{2}{c}{SNLI} & \multicolumn{2}{c}{Multi-NLI} \\
& dev & test & dev & test & dev & test \\
\hline 
\multicolumn{7}{l}{Learning from in domain single task} \\ 
BiLSTM-Max & 87.1 & 86.7 & 84.7 & 84.5 & 70.2/70.8 & 70.8/69.8  \\
\hline
\multicolumn{7}{l}{Learning from 2-datasets and 2-tasks (SNLI and Multi-NLI)} \\ %\hline
Fully Shared & - & - & \underline{85.1} & 85.1 & 71.4/71.2 & 70.8/70.1 \\
Shared-Private & - & - & 85.0 & \underline{\textbf{85.3}} & \underline{\textbf{71.7/71.4}} & \underline{\textbf{71.8/70.6}} \\
Adversarial Shared-Private & - & - & 84.9 & 84.9 & 70.9/71.4 & 71.0/70.0 \\ 
\hline
%\hline
\multicolumn{7}{l}{Learning from 2-datasets and 2-tasks (Quora and SNLI)} \\ %\hline
Fully Shared & 86.8 & 86.6 & 84.9 & 84.8 & - & - \\
Shared-Private & 87.0 & 86.8 & 84.8 & 84.7 & - & -  \\ 
Adversarial Shared-Private & \underline{87.5} & \underline{\textbf{87.0}} & \underline{85.2} & \underline{84.9} & - & -    \\ 
\hline
\multicolumn{7}{l}{Learning from 3-datasets and 2-tasks (Quora and AllNLI)} \\ %\hline
Fully Shared & \underline{87.3} & \underline{86.8} & \underline{\textbf{85.2}} & \underline{84.8} & 69.9/70.3 & 70.0/68.9 \\
Shared-Private & 86.9 & 86.0 & 85.2 & 84.7 & \underline{70.7/70.5} & \underline{70.8/69.3}  \\ 
Adversarial Shared-Private & 87.0 & 86.6 & 84.7 & 84.3 & 70.2/69.4 & 69.6/68.3 \\ 
\hline
%\hline
\multicolumn{7}{l}{Learning from 3-datasets and 3-tasks (Quora, SNLI and Multi-NLI)} \\
Fully Shared & 86.6 & 85.9 & 84.3 & 84.3 & 70.3/69.7 & 70.1/69.6 \\
Shared-Private & \underline{\textbf{87.6}} & \underline{87.0} & \underline{85.2} & \underline{85.2} & \underline{71.2/71.0} & \underline{71.0/70.1}  \\
Adversarial Shared-Private & 86.6 & 86.3 & 84.6 & 84.7 & 70.7/70.7 & 71.0/70.1  \\ 
\hline
\end{tabular}
% }
\caption{Development and test accuracy of source tasks (Quora, SNLI and Multi-NLI) obtained through various multi-task learning architectures. Underlined values indicate best performance among models trained on same set of tasks and bold-faced values indicate best performance among all models. 
}
\label{table:source_task_comparison}
\end{table*}
%%%%%%%%%%%%%%%%%%%%%%%%%%%%%%%%%%%%%%%%%%%%%%%%%%%%%%%%%%%%%%%%

%%%%%%%%%%%%%%%%%%%%%%%%%%%%%%%%%%%%%%%%%%%%%%%%%%%%%%%%%%%%%%%%
\begin{table*}[t]
\centering
\resizebox{\linewidth}{!}{%
\begin{tabular}{@{}l@{\hskip 0.05in}| c@{\hskip 0.15in} c@{\hskip 0.15in} c@{\hskip 0.08in} c@{\hskip 0.1in} c@{\hskip 0.1in} c@{\hskip 0.08in} c@{\hskip 0.08in} c@{\hskip 0.05in} c@{\hskip 0.1in} c@{}}
\hline
Model Type & MR & CR & SUBJ & MPQA & SST & TREC & SICK-R & SICK-E & MRPC & STS14\\ 
\hline 
\hline
\multicolumn{11}{l}{Sentence representation learning from single-task} \\ 
\hline
BiLSTM-Max (on SNLI)    & 80.1	&85.3	&92.6 & 89.1 & \underline{83.6} & \underline{89.2} & 0.885 & 86.0 & 75.2/82.4  &.66/.64\\ 
BiLSTM-Max (on Quora)	&79.2	& 84.6 & 92.6 & 88.8 & 83.5 & 88.0 & 0.861 & 82.4 & 74.8/82.8	& .62/.60 \\
BiLSTM-Max (on Multi-NLI)   & \underline{81.2}	& 85.8	& 93.1	& \underline{89.5}	& 83.4	& 88.8	& 0.863 & 84.7 	& 75.9/83.1  & .66/.63 \\
BiLSTM-Max (on AllNLI)   & 80.9	& \underline{86.3}	& \underline{93.2}	& 89.2	& 83.3	& 88.8 & \underline{0.887} & \underline{86.7} & \underline{76.4/83.4}	&\underline{.69/.66}\\
\hline
\hline
\multicolumn{11}{l}{Sentence representation learning from two-tasks (Quora and SNLI)} \\ \hline
Fully Shared & 79.5 & 84.6 & 92.6 & 89.1 & 82.0 & 88.6 & 0.882 & 85.4 & 74.1/82.3 & \underline{.69/.66} \\
Shared-Private & 80.5 & 84.8 & \underline{93.4} & 89.1 & \underline{84.0} & 90.2 & 0.881 & 86.1 & 75.1/83.2 & .65/.62 \\
Adversarial Shared-Private & \underline{80.9} & \underline{85.4} & \underline{93.4} & \underline{89.2} & 83.6 & \underline{90.8} & \underline{0.886} & \underline{86.9} & \underline{76.5/82.9} & .68/.65 \\
\hline
\hline
\multicolumn{11}{l}{Sentence representation learning from two-tasks (SNLI and Multi-NLI)} \\ \hline
Fully Shared & 81.5	& \textbf{\underline{87.2}} & 92.7 & 89.3 & 84.0 & 89.4 & 0.883 & 86.7	& 75.9/82.8 & .69/.66 \\
Shared-Private & \underline{81.7} &	86.4 & \underline{93.7}	& \underline{89.6}	& \underline{84.8}	& 89.2	& 0.885 & 86.7	& 76.3/82.7 & .67/.64 \\
Adversarial Shared-Private & 81.2 &	86.0 &	93.0 & 89.3	& 83.7 &	\underline{90.4} & \underline{0.886} & \textbf{\underline{87.1}}	& \underline{76.9/83.5}	& \underline{.70/.67} \\
\hline
\hline
\multicolumn{11}{l}{Sentence representation learning from two-tasks (Quora and AllNLI)} \\ \hline
Fully Shared & 81.4 & \underline{86.9} & 93.0 &	\textbf{\underline{89.7}} & \textbf{\underline{85.2}}	& 89.2	& 0.879 & 86.1 & 75.7/82.9 & \textbf{\underline{.71/.67}} \\
Shared-Private & \textbf{\underline{82.0}} & 86.1 & \textbf{\underline{93.9}} & 89.4 & 84.6 & \underline{89.6}	& 0.884 & 86.3 & \underline{76.4/83.4} & .68/.64 \\
Adversarial Shared-Private & 81.4 &	86.3 &	93.2 &	89.4 &	85.1 &	88.4 &	\textbf{\underline{0.888}}	& \underline{86.6}	& 75.5/82.5 & .67/.63 \\
\hline
\hline
\multicolumn{11}{l}{Sentence representation learning from three-tasks (Quora, SNLI and Multi-NLI)} \\ \hline
Fully Shared & 80.9 & 85.4 &	92.5 &	\underline{89.4} &	83.4 &	89.8 &	0.882 & 85.8 & 75.3/82.5 & \underline{.70/.67} \\
Shared-Private & 81.6	&\underline{86.9}	& \textbf{\underline{93.9}}	& 89.2	&\underline{84.4}	& 90.4	& 0.883 & 85.9 	&  76.5/83.3		&.66/.63 \\
Adversarial Shared-Private & \textbf{\underline{82.0}} & 86.3 & 93.8 & \underline{89.4} & 84.1 & \textbf{\underline{92.2}} & \underline{0.884} & \underline{87.0} & \textbf{\underline{77.2/83.6}} & .68/.65 \\
\hline 
\end{tabular}
}
\caption{Transfer test results for various single-task and multi-task learning architectures trained on a combination of Quora, SNLI and Multi-NLI datasets. Underlined values indicates the best performance among models trained on same set of tasks. Bold-faced values indicate the best performance among all models in this table.}
\label{table:comparison}
\end{table*}

%%%%%%%%%%%%%%%%%%%%%%%%%%%%%%%%%%%%%%%%%%%%%%%%%%%%%%%%%%%%%%%%

\begin{table*}[ht]
\centering
\resizebox{\linewidth}{!}{%
\begin{tabular}{@{}l@{\hskip 0.05in}| c@{\hskip 0.15in} c@{\hskip 0.15in} c@{\hskip 0.08in} c@{\hskip 0.1in} c@{\hskip 0.1in} c@{\hskip 0.08in} c@{\hskip 0.08in} c@{\hskip 0.05in} c@{\hskip 0.1in} c@{}}
\hline
Model Type & MR & CR & SUBJ & MPQA & SST & TREC & SICK-R & SICK-E & MRPC & STS14 \\
\hline
\hline
\multicolumn{11}{l}{Shared-Private (trained on DupQuD and SNLI)}  \\ 
\hline
Private Encoder (on DupQuD) & 79.8          & 84.1          & 92.4          & 88.6          & 81.3          & 88.4          & 0.845          & 82.5          & 73.9/81.6          & .60/.58          \\
Private Encoder (on SNLI)  & 79.6          & 84.8          & 92.6          & 89.0          & 82.5          & 87.2          & 0.880          & 85.6          & \underline{75.2/83.2} & \underline{.66/.63} \\
Shared Encoder             & 79.7          & \underline{85.0} & 93.0          & \underline{89.2} & 82.3          & \underline{90.2} & 0.874          & 84.3          & 75.0/83.1          & .65/.62          \\
Combined Encoder           & \underline{80.5} & 84.8          & \underline{93.4} & 89.1          & \underline{84.0} & \underline{90.2} & \underline{0.881} & \underline{86.1} & 75.1/83.2          & .65/.62    \\
\hline
\hline
\multicolumn{11}{l}{Adversarial Shared-Private (trained on DupQuD and SNLI)} \\ 
\hline
Private Encoder (on DupQuD) & 79.5          & 83.1          & 92.1          & 88.6          & 82.1          & 89.0          & 0.860          & 83.4          & 73.9/82.5          & .65/.62          \\
Private Encoder (on SNLI)  & 79.4          & 84.9          & 92.6          & 88.9          & 82.4          & 87.8          & 0.883          & 85.7          & 74.6/82.4          & .67/.65          \\
Shared Encoder             & 80.0          & 84.3          & 92.6          & \underline{89.2} & 81.9          & 86.6          & 0.883          & 85.4          & 73.9/81.5          & \underline{.69/.65} \\
Combined Encoder           & \underline{80.9} & \underline{85.4} & \underline{93.4} & \underline{89.2} & \underline{83.6} & \underline{90.8} & \underline{0.886} & \underline{86.9} & \underline{76.5/82.9} & .68/.65 \\
\hline
\hline
\multicolumn{11}{l}{Shared-Private (trained on SNLI and Multi-NLI)}  \\ 
\hline
Private Encoder (on SNLI) & 79.5 & 84.0 & 92.7 & 89.1 & 82.0 & 87.8 & 0.881 & 84.8 & 75.0/82.7 & .65/.63    \\ 
Private Encoder (on Multi-NLI) & 80.6 & 84.6 & 92.7 & 89.2 & 82.9 & 88.0 & 0.853 & 83.8 & 75.1/82.9 & .60/.58    \\
Shared Encoder & 80.9 & \underline{86.4} & 92.9 & 89.5 & 84.0 & 88.0 & 0.879 & 84.8 & 75.8/83.1 & \underline{.69/.65}    \\
Combined Encoder & \underline{81.7}	& \underline{86.4}	& \underline{93.7}	& \textbf{\underline{89.6}}	& \underline{84.8}	& \underline{89.2}	& \underline{0.885} & \underline{86.7}	& \underline{76.3/82.7}	&.67/.64  \\
\hline
\hline
\multicolumn{11}{l}{Adversarial Shared-Private (trained on SNLI and Multi-NLI)} \\ 
\hline
Private Encoder (on SNLI)     & 79.4 & 84.6 & 92.1 & 89.0 & 82.8  & 86.6  & \underline{0.886}  & 85.5 & 74.0/81.4 & \underline{.68/.66} \\
Private Encoder (on Multi-NLI) & 80.4 & 84.6 & 92.6 & 89.1 & 83.3  & 86.8  & 0.863 & 83.4 & 76.0/83.3 & .65/.63 \\
Shared Encoder                & 81.2     & \underline{86.7}     & 92.4     & 89.3    & 84.5    & 87.0    & 0.875      & 85.1  & 74.8/82.6   & .56/.57        \\
Combined Encoder              & \underline{81.7} & 86.5 & \underline{93.4} & \underline{89.5} & \textbf{\underline{84.9}} & \underline{90.0} & \textbf{\underline{0.888}} & \textbf{\underline{87.1}} & \underline{76.4/83.4} & .64/.63 \\
\hline
\hline
\multicolumn{11}{l}{Shared-Private (trained on DupQuD and AllNLI)}  \\ 
\hline
Private Encoder (on DupQuD) & 79.4 &	82.6 &	92.5 &	88.5 &	82.2 &	89.2 & 0.856	& 82.9	& 74.3/82.4	& .63/.59    \\
Private Encoder (on AllNLI) & 81.1	& \underline{86.1} &	92.9 &	\underline{89.5} &	83.9	&\underline{90.2}	& 0.876	& 85.0 & 76.0/83.3	& .67/.64    \\
Shared Encoder & 81.2 &	85.6 &	93.1 &	89.2 &	83.4	& 88.2	& 0.880	& 85.3 &	75.7/82.8 & \underline{.69/.66}    \\
Combined Encoder&\textbf{\underline{82.0}}	&\underline{86.1}	&\textbf{\underline{93.9}}	&89.4	&\underline{84.6}	&89.6	&\underline{0.884}	&\underline{86.2}	&\underline{76.4/83.4}	&.68/.64\\
\hline
\hline
\multicolumn{11}{l}{Adversarial Shared-Private (trained on DupQuD and AllNLI)} \\ 
\hline
Private Encoder (on DupQuD) & 79.2	& 81.7	& 92.1	& 88.7 & 80.8	& 86.6	& 0.865	& 83.8	& 74.1/82.2	& .67/.64 \\
Private Encoder (on AllNLI) & 81.5	& \underline{86.5}	& 92.8 & 89.4 &	82.9 &	88.4	& 0.885	& 85.5 &	75.7/83.2	& \textbf{\underline{.70/.67}}    \\
Shared Encoder & 80.5	& 84.8	& 92.6	& 89.2 &	\underline{83.2}	& 82.6	& 0.876	& 84.8	& 75.5/83.1 & .57/.56   \\
Combined Encoder & \underline{81.9} &	85.9 & 	\underline{93.0}	& \textbf{\underline{89.6}} &	82.4	& \underline{90.6} &	\underline{0.887}	& \underline{86.7}	& \underline{76.8/83.3} & .61/.60   \\
\hline
\hline
\multicolumn{11}{l}{Shared-Private (trained on DupQuD, SNLI and Multi-NLI)}  \\ 
\hline
Private Encoder (on DupQuD) & 78.9	&83.5	&91.7	&88.3	& 81.4	& 88.6    & 0.850	&82.1	&72.6/81.1	&.62/.59  \\
Private Encoder (on SNLI) &79.1	&83.9	&92.7	&89.0	&81.3	&88.4	   &0.880	&85.2		&74.0/82.0 & .66/.63  \\ 
Private Encoder (on Multi-NLI) &81.0	&85.7	&93.1	&89.3	&82.9	& 89.2	   &0.850	& 84.0		&74.5/82.9  &.60/.58\\ 
Shared Encoder &80.9	&85.8	&92.9	&89.2	&82.7	&85.2	&0.878	&85.8	&76.0/83.1   	& \underline{.68/.65}   \\
Combined Encoder &\underline{81.6}	&\textbf{\underline{86.9}}	&\textbf{\underline{93.9}}	&\underline{89.2}	&\underline{84.4}	& \underline{90.4}	&\underline{0.883}	&\underline{85.9}		&\underline{76.5/83.2}  &.66/.63   \\
\hline
\multicolumn{11}{l}{Adversarial Shared-Private (trained on DupQuD, SNLI and Multi-NLI)} \\ 
\hline
\hline
Private Encoder (on DupQuD)     & 78.9          & 82.3          & 92.2          & 88.8          & 82.3          & 87.2          & 0.855          & 83.0          & 74.3/82.2          & .64/.62          \\
Private Encoder (on SNLI)      & 79.6          & 84.4          & 92.0          & 89.0          & 82.7          & 88.2          & 0.881          & 85.4          & 74.6/82.4          & \underline{.67/.65} \\
Private Encoder (on Multi-NLI) & 80.6          & 84.7          & 93.0          & 89.1          & 83.6          & 89.2          & 0.863          & 84.8          & 75.8/82.8          & .65/.62          \\
Shared Encoder                 & 81.0          & 85.9          & 92.7          & \textbf{\underline{89.6}} & 82.9          & 87.0          & 0.876          & 85.7          & 74.8/82.8          & .66/.64          \\
Combined Encoder               &\textbf{\underline{82.0}} & \underline{86.3} & \underline{93.8} & 89.4          & \underline{84.1} & \textbf{\underline{92.2}} & \underline{0.884} & \underline{87.0} & \textbf{\underline{77.2/83.6}} & .66/.64 \\
\hline
\end{tabular}
}
\caption{Detailed analysis of the transfer test results for shared-private models trained on different combinations of DupQuD, SNLI and Multi-NLI datasets. Combined encoder means the concatenation of shared encoder and all private encoders. Underlined values indicate the best performance among different encoders of the shared-private models trained on the same set of tasks. Bold-faced values indicate the best performance among all models in this table.}
\label{table:sp_comparison}
\end{table*}

% \small
% \bibliography{emnlp2018}
% \bibliographystyle{emnlp_natbib}

% \end{document}
\end{document}